\pdfoutput=1

\documentclass[11pt]{article}

\usepackage[preprint]{acl}

\usepackage{times}
\usepackage{latexsym}

\usepackage[T1]{fontenc}

\usepackage[utf8]{inputenc}

\usepackage{microtype}

\usepackage{inconsolata}

\usepackage{graphicx}

\usepackage{multirow}
\usepackage[most]{tcolorbox}

\usepackage{enumitem}

\usepackage{tabularx}
\usepackage{booktabs}
\usepackage{xcolor}
\usepackage{array}

\title{Why is ``Chicago'' Predictive of Deceptive Reviews? Using LLMs to Discover Language Phenomena from Lexical Cues}

\author{
\textbf{Jiaming Qu\textsuperscript{1}\thanks{This work was done at UNC Chapel Hill and does not reflect the author's position at Amazon.}},
\textbf{Mengtian Guo\textsuperscript{2}},
\textbf{Yue Wang\textsuperscript{2}}\\
\textsuperscript{1}Amazon,
\textsuperscript{2}UNC Chapel Hill\\
qjiaming@amazon.com, mtguo@email.unc.edu, wangyue@email.unc.edu
}

\begin{document}
\maketitle

\begin{abstract}
Deceptive reviews mislead consumers, harm businesses, and undermine trust in online marketplaces. Machine learning classifiers can learn from large amounts of data to distinguish deceptive reviews from genuine ones. However, the distinguishing features learned by these classifiers are often subtle, fragmented, and difficult for humans to interpret, which can hinder user understanding and trust. In this work, we study whether large language models (LLMs) can translate such unintuitive lexical cues into human-understandable language phenomena. We propose a conjecture-then-validate framework, and show that language phenomena obtained in this manner are empirically grounded in data, generalizable across similar domains, and more predictive than phenomena derived from LLMs' prior knowledge or in-context learning. Such phenomena can aid people in critically assessing the credibility of online reviews in environments where deception detection classifiers are unavailable.
\end{abstract}
\section{Introduction}

Online reviews are an essential source of information for consumers to make purchasing decisions and for businesses to understand their customers. Although many platforms have implemented machine learning techniques to detect and block deceptive reviews~\cite{AmazonCoalitionReviews2023}, these systems are not universally available. Users may still face deceptive content without algorithmic assistance. Unlike platforms that can leverage metadata or behavioral signals to identify deception~\cite{paul2021fake}, users rely almost exclusively on review text. Prior work shows that untrained people perform poorly at this task~\cite{bond2006accuracy}.

Therefore, it is important to help users develop the ability to spot potentially deceptive reviews when algorithmic interventions are absent. One possible approach is to expose users to the most salient signals identified by machine learning classifiers~\cite{lime2016,lundberg2017unified,sundararajan2017axiomatic}. However, while classifiers can learn predictive lexical cues from large amounts of data, these cues are often subtle and fragmented. For example, prior work found that the word ``Chicago'' is predictive of deceptive hotel reviews~\cite{lai2020chicago}. Such signals can appear unintuitive and may even reduce users' trust in the machine learning system.

These unintuitive yet predictive words are often not random artifacts. Instead, they reflect underlying language patterns that are not directly observable at the word level. For example, the word ``Chicago'' may reflect a broader pattern where deceptive reviews tend to emphasize brand names or locations. In this work, we view such predictive yet unintuitive lexical cues as surface manifestations of underlying language phenomena. We investigate a key question: \emph{can we translate such lexical cues into explanations that are both human-understandable and grounded in data?} To this end, we develop an LLM-based approach under a conjecture-then-validate framework.

We first prompt LLMs to answer questions in the spirit of ``why is lexical feature $X$ predictive of deceptive or genuine reviews?'' LLMs can generate fluent but potentially unverified explanations, which we refer to as \textbf{conjectures}. These explanations are expressed in natural language and are easy for humans to read. Since plausibility does not guarantee correctness, we then \textbf{evaluate} whether these conjectured phenomena are predictive and generalizable. This evaluation provides evidence that the explanations correspond to real patterns in the data rather than hallucinations.

Our goal is \emph{not} to build a stronger deception detector, but to evaluate whether LLMs can reliably translate machine-learned lexical cues into human-interpretable phenomena. Using deception detection in hotel reviews as a case study, we investigate three research questions (RQs):

\begin{itemize}[leftmargin=*]
    \item \textbf{RQ1 (Predictiveness)}: Can these phenomena distinguish deceptive reviews from genuine ones?
    \item \textbf{RQ2 (Generalization)}: Are these phenomena generalizable to new data in similar domains (i.e., similar products or services)?
    \item \textbf{RQ3 (Alternatives)}: Instead of explaining predictive words, can these phenomena be obtained by prompting LLMs to conjecture based on their prior knowledge or labeled examples?
\end{itemize}

Our results show that LLM-conjectured language phenomena have predictive power, generalize to similar-domain data, and outperform phenomena obtained from LLMs' prior knowledge or in-context learning alone. Our human evaluation further shows that these phenomena improve people's understanding of cues in deceptive reviews. Together, these findings suggest that LLMs can translate model-identified lexical cues into explanations that are both interpretable and empirically supported. These explanations can help users better assess the credibility of online reviews when algorithmic systems are unavailable. 

Our paper makes the following contributions. First, we propose a conjecture-then-validate framework for translating predictive features into human-interpretable language phenomena, and show that these phenomena are empirically grounded and generalizable. Second, we demonstrate through human evaluation that feature importance explanations can identify predictive signals, but they do not explain why these signals are predictive. Our approach takes a step toward bridging this gap by generating explanations that can improve user understanding and calibrated trust in NLP systems.

\section{Related Work}
Feature importance explanations, which highlight input features that are most influential to the model prediction, are a popular approach to explaining machine learning model predictions~\cite{lime2016,lundberg2017unified,sundararajan2017axiomatic}. While empirical studies have shown such explanations can improve end-users' decision-making and understanding of AI systems~\cite{lai2019human,qu2021study}, predictive features are not always self-explanatory. For example, studies that apply such explanations to text data have found that the word ``problems'' is predictive of positive sentiment~\cite{qu2024problems}, and that the word ``Chicago'' is predictive of deceptive reviews~\cite{lai2020chicago}. Given large and representative data, such words are often not the result of overfitting but reflect underlying language phenomena that give rise to these word-label relations. If left unexplained, users may hypothesize wrong reasons for why an unintuitive feature is predictive~\cite{schuff2022human} and even lose trust in a machine learning model~\cite{qu2023,chen2023understanding}. This motivated the need for more interpretable explanations beyond merely showing predictive words.

Prior studies have explored different approaches for making such phenomena more explicit, such as showing nearby words of the predictive word~\cite{ribeiro2018anchors,jacovi2023neighboring,qu2024problems} or considering interactions with other words in the input~\cite{tsang2020does,janizek2021explaining,borisov2022relational}. While these algorithms are effective at revealing context-related phenomena (e.g., negations and colloquial expressions like ``no problems''), they may fail to capture complex phenomena that go beyond local contexts. For instance, the word ``Chicago'' is associated with deceptive hotel reviews because fake reviews often reinforce branding by emphasizing a hotel's full name, including the city name. In this paper, we use deception detection as a case study and leverage LLMs to explain the language phenomena behind deception cues. Our approach was inspired by recent research that prompted LLMs to verbalize predictive features into natural language narratives~\cite{pan2024tagexplainer,zytek2024llms,martens2023tell}. However, these approaches mainly \emph{paraphrase} predictive words for better readability. Our approach is fundamentally different in that it goes beyond observed lexical cues to unobserved phenomena-based explanations.
\section{Methodology}

\subsection{Problem Formulation}
Consider a labeled text dataset $\{(x,y)\}$, where $x$ represents a piece of text and $y$ represents a label (e.g., genuine or deceptive review). Feature importance explanations identify words $\{w \in x\}$ that are predictive of a label $y$. In this work, we focus on a subset $\{w^\prime \in x\} \subseteq \{w \in x\}$: words that are the most salient signals for the classifier but appear unintuitive to humans.
Our goal is to leverage LLMs to translate these salient lexical cues into more human-interpretable language phenomena that plausibly give rise to the cues. Formally, we prompt an LLM to \emph{conjecture} a candidate phenomenon associated with the cue. The LLM-conjectured phenomena may or may not reflect actual patterns in the underlying data. We treat LLM-generated explanations as hypotheses rather than as faithful explanations of model behavior. Thus, we \emph{validate} whether the LLM-conjectured phenomena are predictive of genuine or deceptive reviews (RQ1), generalizable beyond the original data (RQ2), and dependent on classifier-learned predictive words (RQ3). We complement these algorithmic evaluations with human evaluation. 

\subsection{Data Preparation}
\textbf{Dataset}: We used two datasets in our study: a dataset of 800 genuine and 800 deceptive reviews for Chicago hotels~\cite{ott2011finding,ott2013negative} (denoted as $D_{Chicago}$) and another dataset of 240 genuine and 240 deceptive reviews for Houston, New York, and Los Angeles hotels~\cite{li2013identifying} (denoted as $D_{three-cities}$). Genuine hotel reviews were collected from verified travelers while deceptive reviews were written by crowd-sourced workers. 

\textbf{Identifying Predictive Words}: To investigate whether LLMs can translate predictive lexical cues into meaningful language phenomena, we first trained logistic regression classifiers on $D_{Chicago}$ and identified predictive words. It is important to note that we do not treat classification performance as the objective of optimization. We deliberately used a simple model (logistic regression) solely as a \emph{feature-discovery tool}, and predictive words can also be extracted using other sophisticated approaches~\cite{lime2016,sundararajan2017axiomatic,lundberg2017unified}.

Given the limited size of the dataset, we performed 10-fold cross validation with balanced training and testing folds. Each time, we trained a classifier with unigram TF-IDF features, including lowercasing and stop word removal. The classifiers achieved an average F1-score of 0.88. For each classifier, we identified 25 words that were the most predictive of genuine or deceptive reviews through regression coefficients (50 predictive words in total). All selected words passed a Wald test for the significance of predictiveness. To obtain particularly stable lexical cues, we selected only words that were predictive \emph{across all 10 folds}, yielding 16 words for genuine reviews and 14 for deceptive reviews. As shown in Table~\ref{table:phenomena_examples}, we expect that most laypeople without expertise in deception detection would naturally wonder about the underlying language phenomena.

\subsection{Experimental Design}
\textbf{RQ1 Experiment}:
RQ1 investigates whether LLMs provide a reliable approach for discovering language phenomena from lexical cues in deception detection. To this end, we used a conjecture-then-validate pipeline with two separate steps.

First, we prompted the LLM to conjecture the underlying phenomena for words predictive of hotel reviews being genuine or deceptive using the prompt below. We provided all 30 predictive words in the prompt and did not restrict the number of phenomena to be conjectured. All predictive words and their associated phenomena (conjectured by LLMs) are shown in Table~\ref{table:phenomena_examples}. 

\begin{tcolorbox}[
    colback=blue!5,
    colframe=blue!40,
    arc=2mm,
    boxrule=0.8pt,
    left=2mm,
    right=2mm,
    top=2mm,
    bottom=2mm,
    title=\textbf{Prompt for Conjecturing Phenomena}
]
\small
You are an expert in deception detection. You will be provided a list of words that are most predictive of genuine or deceptive Chicago hotel reviews. Your task is to identify language phenomena or psycholinguistic patterns that appear in genuine and deceptive hotel reviews and are related to these words.

\end{tcolorbox}

\begin{table*}[htbp!]
\normalsize
\caption{LLM-conjectured phenomena (aggregated over four LLMs) from predictive words.}
\label{table:phenomena_examples}
\begin{tabularx}{\textwidth}{>{\raggedright\arraybackslash}p{0.4\textwidth} >{\raggedright\arraybackslash}X}
\toprule
\textbf{Predictive Words} & \textbf{LLM-conjectured Phenomena} \\
\midrule

location, street, river, Michigan 
& \textcolor{blue}{Genuine reviews} often provide concrete geographic references and local landmarks. \\

rate, priceline 
& \textcolor{blue}{Genuine reviews} frequently mention pricing, booking platforms, or value for money. \\

reviews, helpful, us 
& \textcolor{blue}{Genuine reviews} refer to other reviews or their desire to contribute useful information. \\

small, large, quiet, bathroom, floor, breakfast 
& \textcolor{blue}{Genuine reviews} use measurable, verifiable adjectives and specific room features. \\

\midrule

luxury, luxurious 
& \textcolor{red}{Deceptive reviews} often overemphasize luxurious aspects and high-end services. \\

hotel, Chicago, millennium 
& \textcolor{red}{Deceptive reviews} often use generic category terms and prominent place or brand names. \\

seemed, recently, definitely 
& \textcolor{red}{Deceptive reviews} tend to use words expressing certainty or vague temporal framing. \\

food, towels 
& \textcolor{red}{Deceptive reviews} list desirable amenities or sensory cues without specific details. \\

husband, vacation, staying, experience 
& \textcolor{red}{Deceptive reviews} use family roles or staged narratives to fabricate a plausible story. \\

\bottomrule
\end{tabularx}
\label{table:phenomena_examples}
\end{table*}

While the conjectured phenomena are plausible, all LLMs are prone to hallucinations. Prior studies have mainly used benchmark datasets~\cite{hendryckstest2021,talmor2019commonsenseqa} or recruited human evaluators~\cite{ziems2024can,goyal2022news} to evaluate LLM-generated contents. However, there is no ground truth for these language phenomena in this task. Therefore, we took an algorithmic evaluation approach: we prompted the same LLM to classify reviews as genuine or deceptive with the conjectured phenomena. The rationale is that if these phenomena reflect actual patterns in the data (i.e., non-hallucinated), they should improve the LLM's predictive performance. Otherwise, fabricated phenomena could mislead the LLM's reasoning. 

Hence, we tested two prompt conditions. First, we use the \emph{phenomena-in-the-prompt} condition, where we provided all the conjectured phenomena associated with both genuine and deceptive reviews in the prompt. Only phenomena but not predictive words were provided. Here, our main objective was to investigate whether LLMs can reliably discover language phenomena at all. Therefore, we prioritized an aggregated evaluation and leveraged the full set of phenomena to test their collective utility. Second, we tested a \emph{zero-shot prompt}, where no auxiliary information was provided.

We tested the pipeline with four LLMs: \texttt{GPT-5-mini} from OpenAI~\cite{OpenAI2025GPT5}, \texttt{Haiku 4.5} from Anthropic~\cite{Anthropic2025ClaudeHaiku}, \texttt{Nova Pro} from Amazon~\cite{AWS2025AmazonNova}, and \texttt{Gemini-2.5-flash} from Google~\cite{DeepMind2025GeminiFlash}. For all LLM calls, we used the corresponding APIs with default settings. Two authors independently designed initial prompts and collaboratively revised the final versions for each prompting scenario. We interacted with LLMs through DSPy~\cite{khattab2023dspy}, a framework for modular construction of LLM applications by explicitly defining prompt components. This design improves the reproducibility of our experiments. Our codebase is available \href{https://jiamingqu.com/deception_detection.zip}{\textcolor{blue}{\underline{here}}}.

\textbf{RQ2 Experiment}: In addition to validating whether the LLM-conjectured phenomena are non-hallucinated, we investigated the generalizability of these phenomena. To this end, we trained logistic regression classifiers on $D_{Chicago}$ and tested on $D_{three-cities}$ using three feature sets: (1) unigram TF-IDF features only, (2) phenomena-based features only, and (3) both features. This setup simulated a scenario where machine learning models are applied to out-of-distribution data: the model learns to differentiate between genuine and deceptive reviews for Chicago hotels and then applies this knowledge to classify hotel reviews from other cities. The rationale is that if these phenomena are generalizable across hotel reviews in different cities, the classifier using phenomena-based features should have better performance than the one using unigram TF-IDF features.

To generate phenomena-based feature values for both training and testing sets, we need to obtain a score $f(r, p)$ for each review $r$ and each phenomenon $p$. Here, $f(r, p)$ measures the extent to which a review $r$ reflects a phenomenon $p$. We approach this problem by computing $P(r|p)$, i.e., the probability that a review $r$ is generated given phenomenon $p$ as the prompt. This can be a proxy for phenomena-based feature values. 

We followed a generative scoring approach using next-token probabilities: $P(r|p) = \prod_{i=1}^{n}{P(w_i|p,w_1, \cdots, w_{i-1})}$, where $p$ is a phenomenon,\footnote{We used the following prompt: \texttt{``Write a hotel review that \{phenomenon\}:''}, where \texttt{\{phenomenon\}} represents one of the nine conjectured phenomena in Table~\ref{table:phenomena_examples}.} and $r$ is a sequence of words $[w_1, \cdots, w_n]$. Each next-word probability $P(w_i|p,w_1, \cdots, w_{i-1})$ was obtained by prompting the LLM with the word sequence $[p,w_1, \cdots, w_{i-1}]$ and reading out the probability of $w_i$ as predicted by the LLM. An open-source LLM that provides a complete probability distribution over \emph{all} possible next words is needed, as $w_i$ might rank at a very low position. We used \texttt{Gemma-7b}~\cite{team2024gemma} for this experiment. We used this approach to generate all nine phenomena-based feature values for reviews in $D_{Chicago}$ and $D_{three-cities}$. It is notable that this approach is one possible approach (and not the only one) to generate such phenomena-based feature values. Even if these approximated feature values are not perfectly accurate, as long as they can improve a deception detection model's performance, it already serves our goal: to show that the phenomena-based features can improve a subsequent deception detector’s out-of-distribution performance. The results in Table~\ref{tab:rq2_results} confirm that this is indeed the case.

\textbf{RQ3 Experiment}: We investigated whether LLMs can effectively discover language phenomena from predictive lexical cues in RQ1. A natural follow-up question is whether these phenomena could also be obtained by prompting LLMs in other ways. Therefore, RQ3 examined whether LLMs can conjecture meaningful phenomena from their prior knowledge or in-context learning. We used \texttt{Haiku 4.5} given its strong performance on deception detection (Table~\ref{tab:rq1_results}) and tested two prompt conditions to conjecture phenomena. In the first condition, we prompted it to conjecture phenomena solely based on its prior knowledge. In the second condition, we randomly sampled 30 genuine and 30 deceptive reviews from $D_{Chicago}$, and prompted it to conjecture phenomena from sampled reviews. We used the same system prompt structure and only changed the input from predictive words to sampled reviews or to nothing. To compare the quality of the phenomena conjectured under each prompt condition (i.e., predictive words, sampled examples, and prior knowledge), we used the same experimental design as in RQ1: we measured \texttt{Haiku 4.5}'s deception detection performance on $D_{Chicago}$ with the conjectured phenomena inserted into the \emph{phenomena-in-the-prompt} condition.

\section{Results}

\subsection{RQ1 Results} 
Table~\ref{tab:rq1_results} shows evaluation results of the four LLMs' predictive performance under different prompt conditions on $D_{Chicago}$. The evaluation results are with respect to the ``deceptive'' label. Across all four LLMs, the \emph{phenomena-in-the-prompt} condition consistently outperformed the \emph{zero-shot} condition. These improvements indicate that \textbf{the conjectured phenomena capture real patterns in the data rather than being entirely hallucinated}.

It is notable that unlike other tasks (e.g., fact-checking and question answering) where ground truth facts exist, there are \emph{no definitive answers} for the conjectured phenomena in this task. Therefore, we define ``hallucinations'' as cases in which LLMs come up with phenomena that are entirely fabricated and not related to any detectable patterns in the data. For instance, a fabricated phenomenon might be ``Deceptive reviews always indicate positive sentiment to promote business''---this is contradicted by the $D_{Chicago}$ corpus where deceptive reviews can still convey negative sentiment. Under this definition, the goal of RQ1 was not to assess whether the LLM-conjectured phenomena represent the exhaustive set of language patterns. Rather, this validity check was to assess whether the LLM-conjectured phenomena are meaningfully associated with real-world examples of genuine or deceptive reviews that any user might see on social platforms. The performance gains LLMs achieved with phenomena in the prompt over the zero-shot prompt indicate that these conjectured phenomena are relevant to the deception detection task.

\begin{table*}[htbp!]
\centering
\caption{RQ1 Results. We evaluated four LLMs' predictive performance on $D_{Chicago}$ under different prompt conditions.}
\begin{tabular}{llllll}
\hline
\textbf{Model} & \textbf{Prompt}      & \textbf{Accuracy} &\textbf{Precision} & \textbf{Recall} & \textbf{F1}    \\ \hline
\multirow{2}{*}{GPT-5 mini} & zero-shot  & 0.6512  & 0.8601  & 0.3612 & 0.5088 \\ 
& with phenomena  & 0.6775   & 0.8272  & 0.4488 & 0.5818 \\ \hline
\multirow{2}{*}{Haiku 4.5} & zero-shot  & 0.6688  & 0.7679  & 0.4838 & 0.5936 \\ 
& with phenomena  & 0.6962   & 0.7532  & 0.5838 & \textbf{0.6577} \\ \hline
\multirow{2}{*}{Nova Pro} & zero-shot  & 0.5512  & 0.6798  & 0.1938 & 0.3016 \\ 
& with phenomena  & 0.5962   & 0.7601  & 0.2812 & 0.4106 \\ \hline
Gemini-2.5 Flash & zero-shot  & 0.6388  & 0.8101  & 0.3625 & 0.5009 \\ 
 & with phenomena  & 0.6600   & 0.8033  & 0.4238 & 0.5548 \\ \hline
\end{tabular}
\label{tab:rq1_results}
\end{table*}

While our goal was \emph{not} to benchmark different LLMs' performance in deception detection or engineer the most effective prompt, these results did reveal two important trends. First, deception detection remained a challenging task for LLMs. Under the zero-shot prompt condition, the four LLMs achieved an average F1-score of 0.4762 (min = 0.3016, max = 0.5936). Since the dataset is balanced, their performance was no better than random guessing. Second, although all LLMs achieved reasonable performance gains with phenomena provided in the prompt, their performance was still worse than that of a logistic regression classifier using unigram TF-IDF features. This traditional model performed surprisingly well on the deception detection task with an F1-score of 0.88.

\subsection{RQ2 Results}
Table~\ref{tab:rq2_results} shows evaluation results of the three logistic regression classifiers' predictive performance. The classifiers used different feature sets and were trained on $D_{Chicago}$ and tested on $D_{three-cities}$. Compared to the classifier using unigram TF-IDF features only, both classifiers using phenomena-based features and combined features achieved better predictive performance on hotel reviews from other cities. These results suggest that \textbf{the LLM-conjectured phenomena captured patterns that are more transferable than unigram lexical cues}.

\begin{table*}[htbp]
\centering
\caption{RQ2 Results. We trained logistic regression classifiers on $D_{Chicago}$ and tested on $D_{three-cities}$ using three different feature sets: unigram only, phenomena only, and both.}
\begin{tabular}{lllll}
\hline
\textbf{Feature Sets}       & \textbf{Accuracy} &\textbf{Precision} & \textbf{Recall} & \textbf{F1}      \\ \hline
unigram only    & 0.7396	& 0.8363	& 0.5958	& 0.6959 \\ \hline
phenomena only & 0.7167	& 0.6688	& 0.8583	& \textbf{0.7518} \\ \hline
unigram+phenomena   & 0.7792	& 0.8602	& 0.6667	& 0.7512 \\ \hline
\end{tabular}
\label{tab:rq2_results}
\end{table*}

This result is expected: while the word ``Chicago'' is the most salient cue of deception in reviews of Chicago hotels, it does not necessarily hold true in hotel reviews from other cities---fabricated review writers might mention hotel names with ``New York'', ``Houston'', and ``Los Angeles'' instead. In fact, regression coefficients of a logistic regression classifier trained on $D_{three-cities}$ corroborated that city names are among the strongest predictors of deceptive hotel reviews within those locales as well. Hence, compared to the unigram text feature ``Chicago'', the corresponding phenomena-based feature---deceptive reviews tend to name-drop hotel and city names---offers a more generalizable insight into deceptive hotel reviews across cities. Compared to lexical features, these language phenomena enable the classifier to learn transferable insights and achieve better predictive performance on out-of-distribution data. Future work could similarly examine whether hotel brand names are a strong signal of deception in other corpora.

\subsection{RQ3 Results}
Table~\ref{table:rq3_results} shows the predictive performance of \texttt{Haiku 4.5} when predicting with phenomena conjectured from three different sources: predictive words, sampled reviews, and prior knowledge. Its performance was highest when phenomena were derived from predictive words, and was noticeably lower when derived from either sampled reviews or prior knowledge alone. These results indicate that \textbf{\texttt{Haiku 4.5} discovered accurate phenomena more effectively when guided by predictive lexical cues than when relying on examples or prior knowledge}. One possible explanation is that predictive words identified by the logistic regression classifier serve as discriminative signals distilled from large amounts of training data. Although these words may appear unintuitive to humans, they anchor the LLM's reasoning toward subtle psycholinguistic patterns that are difficult to infer from a small sample of reviews or from its prior knowledge alone.

\begin{table*}[htbp!]
\centering
\caption{RQ3 Results. We evaluated \texttt{Haiku 4.5}'s predictive performance on $D_{Chicago}$ under the phenomena-in-the-prompt condition with different conjectured phenomena sets: phenomena conjectured from prior knowledge, sampled reviews, and predictive words.}
\label{table:rq3_results}
\begin{tabular}{lllll}
\hline
\textbf{Prompts for conjecturing phenomena}  & \textbf{Accuracy} &\textbf{Precision} & \textbf{Recall} & \textbf{F1}      \\ \hline
prior knowledge  & 0.5988 & 0.6113 & 0.5425 & 0.5748 \\ \hline
sampled reviews  & 0.6269 & 0.6862 & 0.4675 & 0.5561 \\ \hline
predictive words & 0.6962 & 0.7532 & 0.5838 & \textbf{0.6577} \\ \hline
\end{tabular}
\end{table*}

\section{Human Evaluation}\label{sec:human_eval}
\subsection{Study Overview}
A core challenge in evaluating LLM-generated explanations is that there is no ground truth for why a word is predictive of deception or genuineness. While our algorithmic evaluations show that the conjectured phenomena are empirically grounded and generalizable, they do not guarantee that these explanations are meaningful or useful to humans. In particular, explanations may be misleading, under-specified, or difficult to interpret. Moreover, LLM-generated explanations may appear plausible without being faithful to the underlying data. To address these limitations, we complement our algorithmic evaluation with a human-centered evaluation.

We first conducted a large-scale crowdsourced study ($N=220$) where participants performed a deception detection task~\cite{qu2025understanding}. We found that participants who were provided with conjectured phenomena performed significantly better at detecting deception without algorithmic assistance than those who saw predictive words only. To further understand how people interpret predictive lexical cues and the associated language phenomena, we conducted an in-person lab study with a focus on qualitative analysis. The lab study involved 8 English-speaking participants ($N=8$), recruited from our institution. Participants were pre-screened for familiarity with online hotel reviews. The study was approved by our Institutional Review Board (IRB), and participants received a reward of US\$15.

The study followed a within-subjects design with two phases. The task remained the same in both phases. Participants were provided with four genuine and four deceptive Chicago hotel reviews (sampled from $D_{Chicago}$). For each review, participants were shown the predicted label from a logistic regression classifier. Participants were explicitly told that the classifier could make mistakes and should carefully scrutinize each review. The two phases differed only in the explanation provided: (1) \textbf{Lexical cues only}, where predictive words (e.g., ``Chicago'') were highlighted, and (2) \textbf{Lexical cues with phenomena}, where each highlighted word was accompanied by a language phenomenon (e.g., ``Deceptive reviews often use generic category terms and prominent place or brand names''). Participants were instructed to inspect the model prediction and explanation first, and then judge whether the review was genuine or deceptive. After completing both phases, participants took part in a semi-structured exit interview (Appendix~\ref{appendix: interview questions}).

\subsection{User Study Findings}
We audio-recorded and transcribed all interviews conducted over Zoom. One author conducted an inductive thematic analysis on the transcripts. Our analysis focused on how participants interpreted predictive lexical cues, how they used the conjectured phenomena, and how these explanations affected their reasoning and trust in the model. We report four key findings below.

\textbf{From unsupported cues to self-constructed (and often incorrect) explanations.} All participants ($N=8$) reported confusion when only predictive words were shown. However, participants did not simply ignore these cues. Instead, they actively attempted to infer why certain words were highlighted. These interpretations were often speculative and inconsistent across participants. For example, P8 noted, \textit{``I couldn't imagine why those words were highlighted,''}. When these attempts failed, participants either ignored the highlights or formed ad-hoc explanations that were not grounded in the model behavior. This suggests that unexplained lexical cues can lead users to construct their own mental models that might be inaccurate.

\textbf{Language phenomena shifted users from word-level guessing to pattern-level reasoning.} When LLM-conjectured phenomena were introduced, participants changed how they interpreted the model prediction---they began to reason about higher-level language patterns rather than individual words ($N=8$). Participants described the phenomena as providing a ``reason'' behind predictions. Importantly, participants did not passively accept these explanations, but carefully evaluated them. This indicates that the phenomena act as an intermediate representation that connects lexical cues to human-understandable reasoning.

\textbf{Explanations enabled comparison between human and model reasoning.} Most participants ($N=7$) did not uniformly accept the conjectured phenomena. Instead, they selectively accepted explanations that aligned with their own heuristics and rejected those that did not. For instance, P2 agreed with some patterns (e.g., use of extreme superlatives) but disagreed with others. Across participants, we observed that explanations made points of agreement and disagreement more explicit. Rather than persuading users, the phenomena enabled users to compare their own reasoning with the model's reasoning. This indicates that explanation not only improve users' understanding, but also exposing (mis)alignment between human and model perspectives.

\textbf{Explanations supported calibrated trust but did not override prior beliefs.} Participants consistently relied on their own heuristics for deception detection ($N=7$) in the absence of phenomena. The conjectured phenomena did not replace these heuristics. Instead, participants used them as an additional signal. For example, P5 noted that explanations increased confidence only \textit{when the reasoning made sense}. Compared to the lexical-only condition, the phenomena enabled more selective and deliberate reliance on model predictions---users regulate their reliance on the model rather than blindly accepting or rejecting it.
\section{Discussion and Conclusion}

\textbf{Summary of findings}: In this work, we study the feasibility of using LLMs to translate machine-learned lexical cues into higher-level, human-understandable language phenomena in deception detection. We introduce a conjecture-then-validate pipeline: an LLM first \emph{conjectures} the underlying phenomena associated with predictive words, and these phenomena are then \emph{validated} algorithmically by testing whether they reflect patterns in the data.

Our algorithmic results show that these conjectured phenomena have predictive power for distinguishing genuine and deceptive reviews (RQ1), generalize to new, unseen data in the same domain (RQ2), and are most effective when derived from predictive words rather than prior knowledge or in-context learning (RQ3). Together, these findings suggest that LLMs can translate nuanced lexical cues into human-interpretable language phenomena in deception detection---a capability that may extend to other text classification tasks where predictive features are unintuitive but reflect underlying linguistic patterns.

Our human evaluation complements these findings by showing how such phenomena affect user reasoning. When only predictive words were shown, participants attempted to infer their meanings but often formed speculative or inconsistent interpretations. In contrast, when language phenomena were provided, participants shifted from reasoning about isolated words to reasoning about higher-level patterns. Importantly, participants did not passively accept these explanations. Instead, they compared them with their own heuristics and selectively accepted or rejected them. As a result, the conjectured phenomena did not lead to blind trust, but supported calibrated trust.

\textbf{Implications for human trust in NLP systems}: While feature importance explanations are a common approach to explaining model predictions, we find that predictive words do not always make immediate sense to humans. LLMs can support model interpretability by discovering intermediate, human-interpretable representations that connect machine-learned features with human reasoning. Building on salient lexical cues, these representations provide a structured way for users to interpret model outputs and relate them to higher-level language patterns. This highlights a key tension between plausibility and faithfulness in generating human-centered explanations for NLP systems: user trust should not be achieved by making explanations more persuasive, but by enabling users to assess model outputs in relation to their own knowledge. In our study, LLM-conjectured phenomena supported this process by improving interpretability while preserving user agency. Users remained active decision-makers who could accept, reject, or question the model’s reasoning. More broadly, translating unintuitive predictive features into meaningful language phenomena can improve how users interact with NLP systems and support appropriate reliance rather than passive acceptance.

\textbf{Future work}: While our validation approach provides evidence that the LLM-conjectured phenomena correspond to real patterns in the data, it does not guarantee that these explanations are fully faithful to the underlying model. Some phenomena may reflect simplified interpretations rather than complete causal mechanisms. Future work should investigate how to better assess and improve the faithfulness of LLM-generated explanations. Additionally, our human evaluation is limited in scale and domain, and future studies could examine broader user populations and other NLP tasks.
\section{Limitations}
This work has several limitations. First, although our conjecture-then-validate framework provides evidence that LLM-conjectured phenomena correspond to real patterns in the data, it does not guarantee full faithfulness to the underlying model or causal mechanisms. Some explanations may still reflect partial or simplified interpretations.

Second, our approach relies on single-pass prompting without iterative refinement. Future work could explore human-in-the-loop or multi-step strategies to improve the quality and reliability of the generated phenomena.

Finally, we use off-the-shelf LLMs and focus on deception detection in hotel reviews, which may limit generalizability. Our human evaluation also primarily involves lay participants; expert linguistic evaluation could provide deeper validation. 
\section{Ethical Considerations and Broader Impacts}

This study was approved by our Institutional Review Board (IRB). Participants were informed about the task and compensated appropriately. The datasets used are publicly available or previously collected, and no personally identifiable information was used. Participants in our study were explicitly informed that the reviews were pre-labeled and part of a research setting.

Our work aims to improve human understanding and support calibrated trust in NLP systems. However, the identified language phenomena could be misused to craft more sophisticated deceptive content. Such misuse could make deceptive content more difficult to detect by both humans and automated systems. More broadly, LLM-generated explanations may appear plausible without being fully faithful to underlying data or models. If deployed without proper validation, such explanations could mislead users or create a false sense of understanding. We emphasize that our framework includes an explicit algorithmic validation step to mitigate this risk, and we encourage future work to further investigate methods for ensuring faithfulness and preventing misuse.

\bibliography{custom}

\newpage
\appendix



\section{Exit Interview Questions}\label{appendix: interview questions}
Participants were asked the following questions in the exit interview.

\begin{enumerate}
    \item The AI system suggests its predictions and highlights words that are strong evidence for both labels. For any of the reviews, did you find yourself looking at those highlighted words? Were there any particular goals you were hoping to achieve by looking at these words?
    \item Did you find these highlighted words helpful to you when understanding the AI's prediction of whether a review was genuine or deceptive? If yes, how did they help? If no, why were they not helpful? 
    \item The AI system suggests its predictions and highlights words that are strong evidence for both labels. It also presents some language phenomena associated with predictive words. For any of the reviews, did you find yourself looking at those language phenomena? Were there any particular goals you were hoping to achieve by looking at these phenomena? 
    \item Did you find these phenomena helpful to you when understanding the AI's prediction of whether a review was genuine or deceptive? If no, why were they not helpful? 
\end{enumerate}

\end{document}